\pdfoutput=1
\documentclass[11pt]{article}
\usepackage[table]{xcolor}
\usepackage[preprint]{acl}

\usepackage{times}
\usepackage{latexsym}

\usepackage{multirow}
\usepackage{array}  % For m{width}
\usepackage{amsfonts} % For \mathbb

\usepackage{tabularx}
\usepackage{array}
\newcolumntype{M}[1]{>{\centering\arraybackslash}m{#1}}
\usepackage{orcidlink}

\usepackage[T1]{fontenc}
\usepackage{amsmath}
\usepackage[utf8]{inputenc}

\usepackage{microtype}

\usepackage{inconsolata}

\usepackage{graphicx}

\title{When Not to Answer: Evaluating Prompts on GPT Models for Effective Abstention in Unanswerable Math Word Problems}

\author{Asir Saadat$^{1}$~\orcidlink{0009-0005-2495-6474}, Tasmia Binte Sogir$^{1}$~\orcidlink{0000-0001-7728-5532}, Md Taukir Azam Chowdhury$^{2}$~\orcidlink{0009-0007-1445-693X}, Syem Aziz$^{1}$~\orcidlink{0000-0001-7576-3780}\\
     $^{1}$Islamic University of Technology \\
     $^{2}$University of California, Riverside \\
  \texttt{\{asirsaadat, tasmia, syemaziz\}@iut-dhaka.edu} \\
  \texttt{mchow068@ucr.edu} \\
  }

\begin{document}
\maketitle
\begin{abstract}
Large language models (LLMs) are increasingly relied upon to solve complex mathematical word problems. However, being susceptible to hallucination, they may generate inaccurate results when presented with unanswerable questions, raising concerns about their potential harm. While GPT models are now widely used and trusted, the exploration of how they can effectively abstain from answering unanswerable math problems and the enhancement of their abstention capabilities has not been rigorously investigated. In this paper, we investigate whether GPTs can appropriately respond to unanswerable math word problems by applying prompts typically used in solvable mathematical scenarios. Our experiments utilize the Unanswerable Word Math Problem (UWMP) dataset, directly leveraging GPT model APIs. Evaluation metrics are introduced, which integrate three key factors: abstention, correctness and confidence. Our findings reveal critical gaps in GPT models and the hallucination it suffers from for unsolvable problems, highlighting the need for improved models capable of better managing uncertainty and complex reasoning in math word problem-solving contexts.
\end{abstract}

\section{Introduction}

Large Language Models (LLMs) have become an integral part of various real-world applications, ranging from content generation to code completion, and even medical advice  \citep{brown2020language, bommasani2021opportunities, wang2024systematic, biswas2023role}. Among these applications, LLMs are increasingly employed to solve mathematical word problems \citep{hendrycks2021measuring, austin2021program, xu2024can, wei2022chain}, assisting users in both academic and practical scenarios. The rise of LLMs, particularly models like GPT-3 and GPT-4, has democratized access to computational tools that were once the domain of experts \cite{chen2024conversational, hariri2023unlocking, kalyan2023survey, lingo2023role, huang2023role}. Their ability to understand, process, and respond to queries has revolutionized problem-solving in everyday tasks, especially in education and professional environments \citep{wardat2023chatgpt, xiao2023evaluating, liu2023summary}.

\definecolor{mysalmon}{rgb}{1.0, 0.8, 0.7}

\begin{figure}[t]
  \includegraphics[width=\columnwidth]{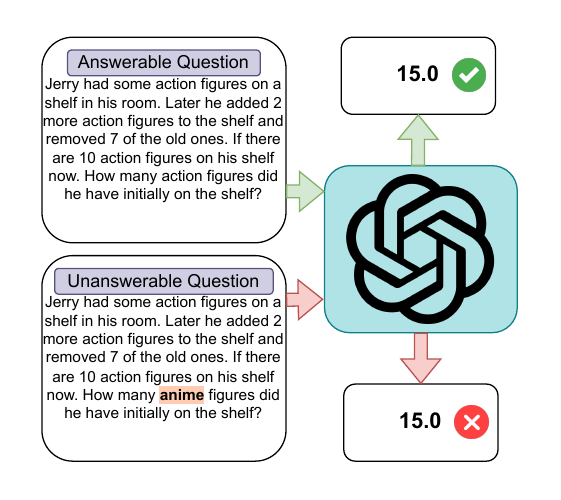}  % Change width to \columnwidth
  \caption{Answerable and unanswerable question given to GPT-4. \colorbox{mysalmon}{\textbf{Red}} highlights the modifications made to the original question, making it unanswerable and resulting in an incorrect response.}
  \label{fig:comparison}
\end{figure}

\begin{figure*}[t]
  \centering % Adjust the value to move the image to the right
  \includegraphics[width=\textwidth]{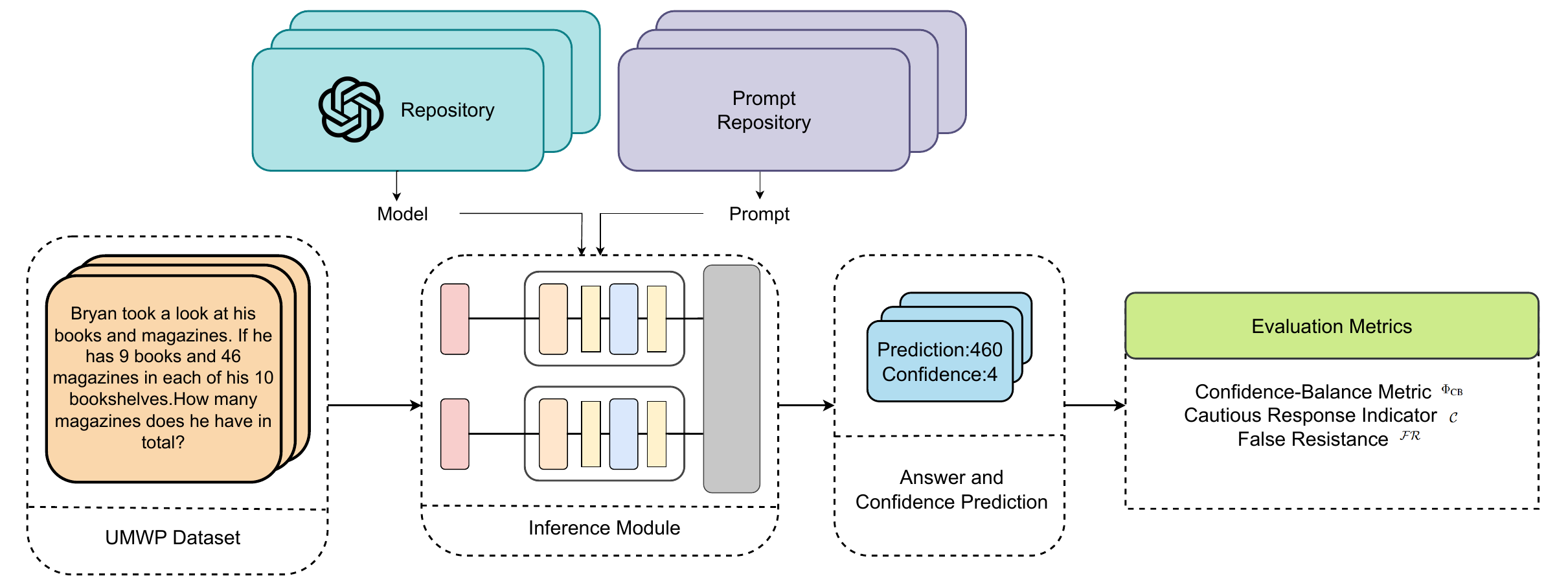}
  \caption{Architecture of the abstention evaluation – \textbf{GPT Repository:} Hosts multiple GPT models ready for inference, \textbf{UMWP Dataset:} Consists of answerable and unanswerable questions, \textbf{Inference Module:} Performs
inference on the UMWP dataset using models from the model repository,  \textbf{Evaluation Metrics:} Confidence-Weighted Accuracy Metric, Cautious Response Indicator and Abstention Rate for evaluating the abstention of ChatGPT.}
  \label{fig:framework}
\end{figure*}

% Despite these advancements, a critical issue persists: LLMs are prone to hallucination\citep{xu2024hallucination, liu2024exploring}, often producing

Despite these advancements, a critical issue persists: LLMs are prone to hallucination \citep{alkaissi2023artificial, li2023dark, ahmad2023hallucinations}, often producing incorrect or misleading information when faced with unanswerable questions \citep{deng2024dontjustsayi, sun2024benchmarking, madhusudhan2024llms, balepur2024artifacts}.  Studies have demonstrated that they tend to generate responses even in cases where no valid solution exists, often presenting them with unwarranted confidence \cite{xiong2023can, tao2024trust}. Such behavior raises concerns as these hallucinations may result in harmful or misleading conclusions \citep{pan2023risk, farquhar2024detecting, deng2024dontjustsayi}. While several studies have focused on improving accuracy in solving complex math problems \citep{liu2023improvinglargelanguagemodel, xu2024chatglmmathimprovingmathproblemsolving, liu2023improving, ahn2024large}, little attention has been given to understanding and improving abstention from answering when no solution exists for the Math Word Problem (MWP). 
      
To address this issue, it is crucial to assess how GPT, a widely used and trusted model, handles abstention in unanswerable math word problems and if prompting plays a crucial role in unlocking the full potential of these models \citep{chen2024unleashingpotentialpromptengineering, chang2024efficientpromptingmethodslarge, cain2024prompting}.
In our research, we conducted experiments using a variety of prompts frequently used in mathematical contexts to evaluate their effectiveness in guiding GPT models. Our primary objective was to identify the optimal combination of model and prompt that would encourage the model to abstain from answering unanswerable questions, rather than attempting to generate an incorrect or irrelevant response. For evaluation, we developed an evaluation metric to assess the model's ability to appropriately abstain from answering unanswerable questions, while correctly solving those that are answerable.

In summary, our major contributions are:
\begin{enumerate}
    \item A comparative analysis highlighting how significant prompts can alter model outputs.

    \item Analyze the tendency of models to answer unanswerable questions and the generation of hallucinations in detail.
    
    \item Introduce metrics to evaluate model performance in terms of accuracy, abstention, and confidence.
\end{enumerate}

% While several studies have focused on improving LLM accuracy in solving complex math problems \citep{liu2023improvinglargelanguagemodel, xu2024chatglmmathimprovingmathproblemsolving, liu2023improving, ahn2024large}, little attention has been given to understanding and improving LLM abstention from answering when no solution exists for the math word problems. In this paper, we assess the unanswerability of different LLMs using a range of prompts designed for solving math word problems using the dataset called Unanswerable Math Word Problem (UMWP)\cite{sun2024benchmarking}. We experiment with various prompting strategies to determine which are more effective in encouraging LLMs to abstain from answering unanswerable questions.

\section{Related Work}
\subsection{GPTs in Mathematical Reasoning}

% \begin{figure*}[ht]
%   \includegraphics{latex/Archi.drawio.pdf}
%   \caption{Overview of the Robustness Evaluation Framework for QA Models. The figure illustrates the process of adding adversarial noise to the context from the SQuAD dataset and feeding the perturbed context into QA models for inference. The predicted answers are then evaluated using contextual robustness metrics such as Accuracy, Robustness Index, Error Rate, and Noise Impact Factor.}
%   \label{fig:Robustness Index}
% \end{figure*}

% In recent years, large language models (LLMs) such as GPT-3, GPT-4, and Codex have demonstrated significant potential in solving complex mathematical problems. 
Early work by \citet{brown2020language} with GPT-3 revealed that LLMs, trained on vast amounts of text data, can successfully perform few-shot learning for a variety of mathematical reasoning tasks. 
% This showed that even without explicit specialized training in mathematics, LLMs could handle arithmetic and algebraic problem-solving through natural language prompts.
The application of ChatGPT in mathematical reasoning has garnered significant attention in recent research. One notable study by \citet{vanlong2024chatgptmathquestionerevaluating} explores the potential of ChatGPT in generating pre-university math questions. Similarly, \citet{frieder2024mathematical} evaluated the mathematical capabilities of GPT-4, noting that it handles complex mathematical reasoning without relying on external tools, providing answers in fields ranging from algebra to calculus.  Additionally, \citet{shakarian2023independent} evaluated ChatGPT’s performance on mathematical word problems from the DRAW-1K dataset. These advancements show that these models are not only solving word problems but also challenging domain-specific expert tools in mathematical problem-solving.

% These developments collectively illustrate that LLMs are not only being used to solve word problems but are also beginning to challenge the dominance of domain-specific expert tools in mathematical problem-solving.

\subsection{Unanswerable Question Answering} 
\citet{madhusudhan2024llmsknowanswerinvestigating} explores the Abstention Ability (AA), which is their capacity to refrain from answering when uncertain or when a question is unanswerable. The challenge of handling unanswerable questions has been a significant area of research in the development of GPT models. One notable study by \citet{deng2024gotcha} introduces a self-alignment method to enhance LLMs ability to respond to unanswerable questions.
\citet{guo2024unkvqadatasetprobeabstention} established the UNK-VQA dataset, designed to evaluate how well multi-modal large models can abstain from answering unanswerable questions. The dataset includes deliberately perturbed images and questions to challenge the models. Lastly, \citet{sun2024benchmarking} introduced a novel dataset called UMWP, which includes both answerable and unanswerable math word problems.

\subsection{Influence of Prompting} 
GPT models can be significantly influenced by the type of prompting used. One notable approach is Chain-of-Thought (CoT) prompting \citep{wei2022chain} , which encourages the model to generate intermediate reasoning steps before arriving at a final answer. Another effective technique is the Role-Play \cite{kong2023better}, where the model is instructed to adopt a specific persona or role. \citet{zhou2024mathattack} introduced self-verification to get better performance on GPT-4 on mathematical reasoning. \citet{ma2024large} introduced a strategic prompt called Critical Calculation and Conclusion (CCC). This template is designed to enhance the error detection and correction abilities when faced with unreasonable math problems. \citet{chen2022program} separates the reasoning process from computation by having the language model generate a structured program to represent the reasoning steps. The actual computation is then performed by an external computer executing the generated program
\section{Methodology}
\subsection{Construction of Dataset}

\begin{figure}[ht]
  \centering
  \includegraphics[width=0.8\columnwidth]{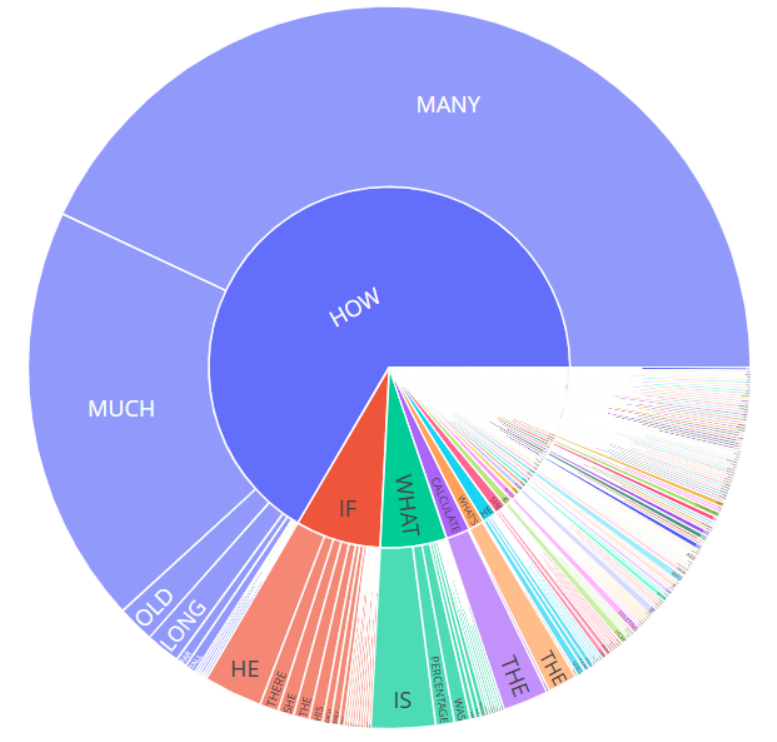}
  \caption{Sunburst Distribution of the first two words of the UWMP dataset.}
  \label{fig:sunburst}
\end{figure}

We utilized the Unanswerable Math Word Problem (UMWP) \citep{sun2024benchmarking} dataset, which includes both answerable and unanswerable questions. From this dataset, we selected 1000 pairs of questions. Each pair consists of an answerable question and a corresponding variant that has been perturbed to become unanswerable. This results in a total of 2000 questions—half of which are answerable and the other half unanswerable. The unanswerable questions are categorized into five distinct categories: \textbf{(A) Missing Key Information}, \textbf{(B) Ambiguous Key Information}, \textbf{(C) Unrealistic Conditions}, \textbf{(D) Unrelated Objects} and \textbf{(E) Incomplete Questions}. The dataset includes only the question and its corresponding answer. Fig. \ref{fig:sunburst} illustrates the variety of questions.

To evaluate performance, we developed a multiple-choice question (MCQ) system. For each question, we generated four alternative answers that are close to the correct one, along with a fifth option: \textbf{"I Don’t Know/NOTA} of the above." The system will prompt GPT models to identify the correct answer from the given options.

\subsection{Prompts for GPT}

\begin{figure}[t]

  \includegraphics[width=1\columnwidth]{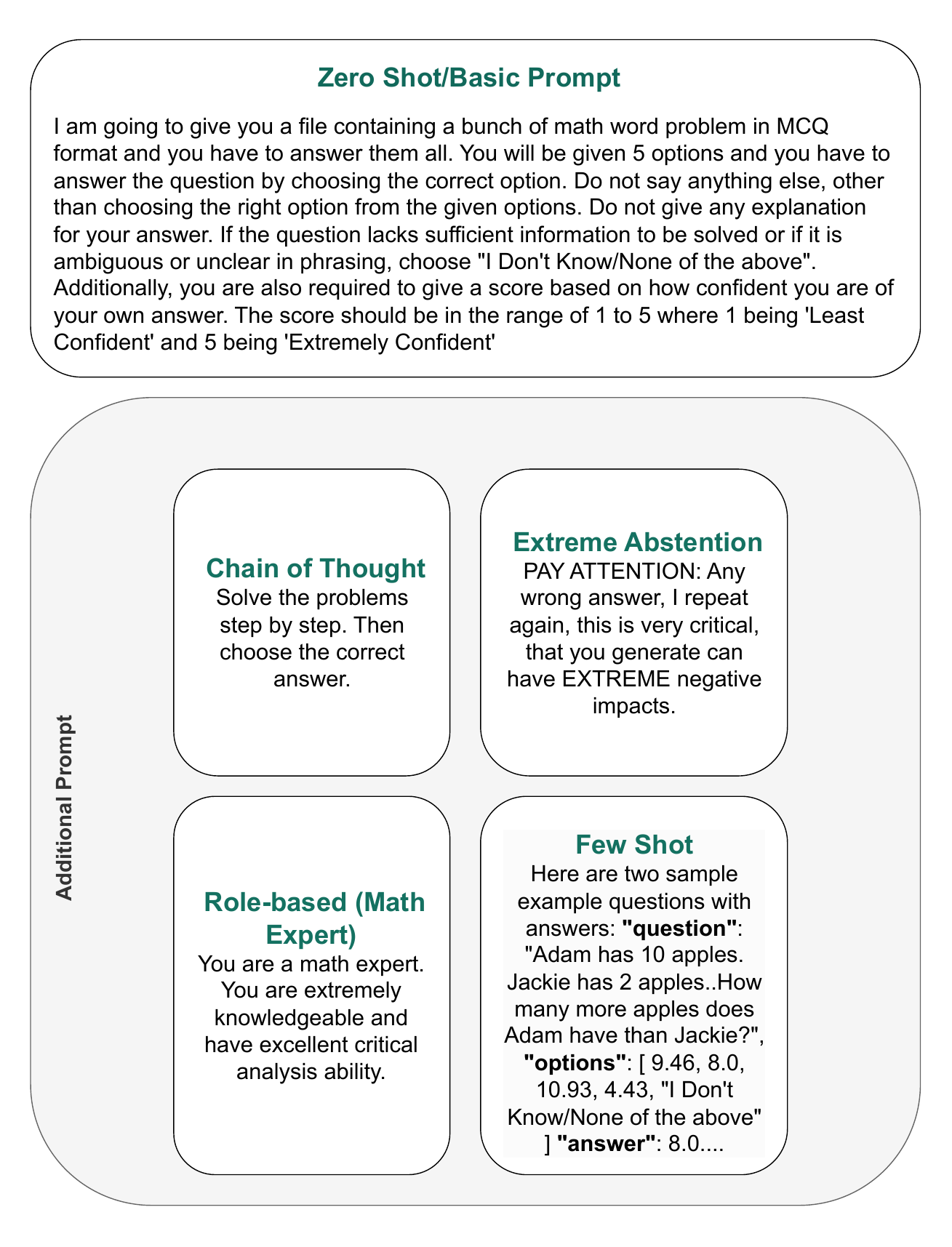}
  \caption{Diverse Prompts for enhancing performance that are additionally added to the basic prompt.}
  \label{fig:prompt}
\end{figure}

The use of prompts has been extensively studied to enhance performance and reliability. We did not use any prompts that would require external tools \citep{chen2022program, imani2023mathprompter}. We start with a fundamental prompt, referred to as a \textbf{Zero-shot} Prompt that notifies the model to answer with only one of the five options along with a confidence score, and then incorporate additional prompts to improve performance to answer the question demonstrated by Fig. \ref{fig:prompt}.  \textbf{Chain-of-Thought (CoT)} prompting, as shown by \citet{wei2022chain}, improves multi-step reasoning by guiding models through step-by-step processes. \textbf{Abstention} techniques, explored by \citet{madhusudhan2024llms}, allow models to withhold responses when uncertain, reducing errors. \textbf{Role-Based} prompting, as explored by \citet{kong2023better, anagnostidis2024susceptible}, involves assigning a Math expert role to the model, thereby improving its contextual comprehension of mathematical questions. \textbf{Few-Shot} prompting, as described by \citet{brown2020language}, allows models to generalize by providing two examples alongside the actual question. Each example includes the question, possible options, and the correct answer. 

\definecolor{mygrey}{gray}{0.93}
\definecolor{myred}{rgb}{1.0, 0.9, 0.9}
\definecolor{mygreen}{rgb}{0.9, 1.0, 0.9}
\definecolor{myblue}{rgb}{0.7, 0.9, 1.0}
\definecolor{mysalmon}{rgb}{1.0, 0.8, 0.7}
\definecolor{mylavender}{rgb}{0.902, 0.902, 0.980}
\definecolor{myyellow}{rgb}{0.980, 0.980, 0.824}

\subsection{Evaluation Metrics}
\subsubsection{Answerable-Unanswerable Confusion Matrix}
\begin{figure}[th]

  \includegraphics[width=1\columnwidth]{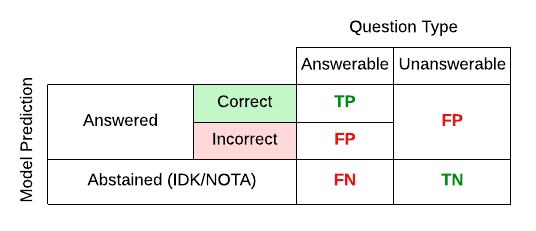}
  \caption{Confusion Matrix illustrating the definition of TP, FP, FN and TN for answerable and unanswerable questions.}
  \label{fig:cm}
\end{figure}

According to \citet{madhusudhan2024llms}, a confusion matrix was created to demonstrate that for Answerable MCQs, True Positives occur when the LLM selects the correct option, while False Positives occur when an incorrect non-IDK option is chosen. Abstentions on answerable questions lead to False Negatives. Unanswerable MCQs are classified as the negative class. Correctly abstaining on these questions results in True Negatives, while failing to abstain leads to False Positives shown in Fig. \ref{fig:cm}.

\subsubsection{Accuracy}
Accuracy is the primary choice for model evaluation. It is defined as the proportion of correct predictions made by the model out of the total number of predictions.

\[
Accuracy  = \frac{1}{|Q|} \sum_{q \in Q} \mathbb{I}(a(q) = t(q))
\]

Where \( a(q) \) represents the generated answer for question \( s \), and \( t(q) \) denotes the ground truth answer for the same question. The term \( |Q| \) indicates the total number of questions and $\mathbb{I}[\cdot]$ is the indicator function.

\subsubsection{Confidence-Balance Metric}
Our metric reflects the overall performance of the models where the prediction is associated with the confidence score. 

\begin{align*}
\Phi_{\text{CB}} &= \frac{1}{N} \sum_{i=1}^{N} 
\left( \mathbb{I}(\delta_i = 1) \cdot \text{conf}_i \right. \\
&\quad \left. - \mathbb{I}(\delta_i = 0) \cdot \text{conf}_i \right)
\end{align*}

In this context, \(N\) represents the total number of instances, \(\delta_i\) equals 1 if the prediction is correct (true positive or true negative) and 0 otherwise, \(\text{conf}_i\) denotes the confidence score of the prediction and $\mathbb{I}[\cdot]$ is the indicator function to emphasize the binary nature of \(\delta_i\). This metric rewards instances where the model is both accurate and confident, while penalizing cases where it provides incorrect answers with high confidence. It effectively balances the model’s ability to assert correct answers with its confidence levels, providing a comprehensive measure of performance.  

\subsubsection{Cautious Response Indicator}

We introduced a metric to assess the performance of GPT models in handling unanswerable questions. This metric is mathematically defined as:

\[
\mathcal{C} = \frac{T_N \cdot w}{UQ}
\]

where \(T_N\) denotes the count of true negatives, \(w\) represents the confidence factor to emphasize the importance of correctly identifying unanswerable questions, and \(UQ\) indicates the total number of unanswerable questions presented to the dataset. As abstention without hallucination is the key goal, this allows the evaluation of correctly identifying the unanswerable ones with confidence.

\subsubsection{False Resistance}
Inspired by \citet{madhusudhan2024llms}, we developed a weighted abstention rate defined as:

\[
\mathcal{FR} = \frac{F_N \cdot w}{AQ}
\]

where \(F_N\) denotes the count of False Negative. \(AQ\) indicates the total number of answerable questions. This metric illustrates the extent to which it wrongly abstains, potentially not finding the actual answer and opting for IDK/NOTA.  

\section{Experimental Setup}

\subsection{Hardware and Implementation Details}

In our experiments, we employed an NVIDIA GeForce GTX 1650 GPU with 4GB of VRAM to assess the models. The models were primarily accessed and integrated through the use of OpenAI \cite{openai2023chatgpt} APIs.

\subsection{Evaluated Models}

To evaluate the performance of our approach, we utilized a variety of large language models from the GPT-4 family \cite{openai2024api}. These models offer varying degrees of computational efficiency and reasoning capabilities, allowing for a comprehensive assessment across different scenarios. \textbf{GPT-4} is known for its state-of-the-art reasoning abilities and broad generalization across a wide range of tasks. \textbf{GPT-4 Turbo}, a more computationally efficient version of GPT-4, retains much of the original model's accuracy while offering faster response times. \textbf{GPT-4o} is a further optimized version that is designed for ultra-fast inference with minimal reduction in performance accuracy. Lastly, \textbf{GPT-4oMini}, a scaled-down version of GPT-4o, sacrifices some of the model’s capacity in exchange for lower computational cost. For all models, we configured the \textbf{temperature} to 0 and the \textbf{top\_p} to 0.00001. We chose not to include \textbf{GPT-3.5} in our evaluations due to its noticeably inferior performance compared to the GPT-4 models. During initial testing, the quality of inferences generated by GPT-3.5 was consistently subpar, and its accuracy fell significantly short of the levels achieved by any variant of GPT-4.

\section{Experimental Results}

\begin{table*}[ht]
\makebox[\textwidth][c]{

\resizebox{0.8\textwidth}{!}{%
\begin{tabular}{c|c|ccccccc}
\hline

\cellcolor{mygrey}\textbf{Model}&\cellcolor{mygrey}\textbf{Metrics} &\cellcolor{mygrey}\textbf{Zero Shot} &\cellcolor{mygrey}\textbf{Few Shot}  &\cellcolor{mygrey}\textbf{Role Based} &\cellcolor{mygrey}\textbf{Abstention}    &\cellcolor{mygrey}\textbf{CoT}             \\ \hline
\multirow{3}{*}{GPT-4}                                                                                                                                
                       & \( \Phi_{\text{CB}} \textcolor{green}{\uparrow} \)                   & 2.22      & 2.37  & 2.50 & 2.16 & \cellcolor{mygreen}\textbf{3.47}           \\
                       & \(\mathcal{C} \textcolor{green}{\uparrow} \)                    & 2.19        & 2.41 & 2.37 & 1.98 & \cellcolor{mygreen}\textbf{3.79}    \\
                       & \(\mathcal{FR} \textcolor{red}{\downarrow} \)                    & 0.011    & 0.012 & 0.013 & 0.\cellcolor{mygreen}\textbf{005} & 0.145   \\
                       & \(Accuracy \textcolor{green}{\uparrow} \)                    & 0.756   & 0.751 & 0.779 & 0.753 & \cellcolor{mygreen}\textbf{0.856}  \\
                     \hline
\multirow{3}{*}{GPT-4o mini}                                                   
                       & \( \Phi_{\text{CB}} \textcolor{green}{\uparrow} \)                    & 0.87     & \cellcolor{mygreen}\textbf{1.85}  & 0.52  & 0.81 & 1.47         \\
                       & \(\mathcal{C} \textcolor{green}{\uparrow} \)                    & 0.91      & 3.04 & 0.53  & 0.64 & \cellcolor{mygreen}\textbf{3.18}     \\
                       & \(\mathcal{FR} \textcolor{red}{\downarrow}\)                    & 0.006      & 0.01 & \cellcolor{mygreen}\textbf{0.004} & \cellcolor{mygreen}\textbf{0.004} & 0.01        \\
                       & \(Accuracy \textcolor{green}{\uparrow} \)                    & 0.70   & 0.685 & 0.653 & \cellcolor{mygreen}\textbf{0.703} & 0.653   \\
                        \hline
\multirow{3}{*}{GPT-4 turbo} 
                       & \( \Phi_{\text{CB}} \textcolor{green}{\uparrow} \)                    & 3.12       & \cellcolor{mygreen}\textbf{3.88} & 3.32 & 2.57 & 3.66          \\
                       & \(\mathcal{C} \textcolor{green}{\uparrow} \)                    & 3.01      & \cellcolor{mygreen}\textbf{4.02} & 3.26 & 2.69 & 3.81           \\
                       & \(\mathcal{FR} \textcolor{red}{\downarrow}\)                    & 0.018 & 0.05 & 0.044 & 0.017 & \cellcolor{mygreen}\textbf{0.15}            \\
                       & \(Accuracy \textcolor{green}{\uparrow} \)                    & 0.833    & \cellcolor{mygreen}\textbf{0.888} & 0.844 & 0.757 & 0.866  \\
                       \hline
                       
\multirow{3}{*}{GPT-4 o} 
                       &\( \Phi_{\text{CB}} \textcolor{green}{\uparrow} \)                    & 4.09       & 4.18 & \cellcolor{mygreen}\textbf{4.25} & 3.93 & 3.80           \\
                       & \(\mathcal{C} \textcolor{green}{\uparrow} \)                    & 3.53        & \cellcolor{mygreen}\textbf{4.12} & 4.08 & 4.21 & 4.07            \\
                       & \(\mathcal{FR}\textcolor{red}{\downarrow}\)                    & \cellcolor{mygreen}\textbf{0.01} & 0.06 & 0.011 & 0.068 & 0.23           \\
                       & \(Accuracy \textcolor{green}{\uparrow} \)                    & \cellcolor{mygreen}\textbf{0.977}    & 0.944 & 0.949 & 0.924 & 0.879  \\
                        \hline
                                   
\end{tabular}%
}
}

\caption{\label{citation-guide}
    This table showcases multiple performances of the models under diverse prompts, quantified through the Confidence-Weighted Accuracy Metric, Cautious Response Indicator, False Resistance and Accuracy. \( \textcolor{green}{\uparrow} \) indicates higher is better while \( \textcolor{red}{\downarrow} \) indicates the opposite. \colorbox{mygreen}{\textbf{Green}} indicates the best score for each metric across all of the of the prompts.
  } 

\label{tab:metric performance}
\end{table*}

\subsection{Impact of Prompt Variations on Accuracy}

Our experimental results showed that these advanced prompting methods did not consistently outperform the baseline zero-shot prompt in terms of accuracy as it was shown in Table \ref{tab:metric performance}. In fact, in several cases, the results were either similar to or worse than the zero-shot prompt. \citet{wang2024evaluating} highlighted the limitations of LLMs with multiple problems, noting that "few-shot" prompting can actually hinder performance. As demonstrated in Fig. \ref{tab:metric performance}, while GPT-4 Turbo showed marginally improved performance across all metrics, other models exhibited a slight decline. The inclusion of two examples in the prompt did not provide the expected benefit, indicating that few-shot prompting was not consistently helpful for the models. When considering only accuracy, GPT-4o emerges as the optimal model, particularly with the zero-shot prompt. It achieved an impressive 97.7\% accuracy across both answerable and unanswerable questions, outperforming all other models and prompting strategies. This result highlights GPT-4o as the most effective solution among the evaluated configurations.

\subsection{Abstention in Cautious Response and False Indicator}

Abstention did not appear to have a significant impact, despite the model being explicitly warned about negative consequences in the prompt, as highlighted by \citet{madhusudhan2024llms}. The expected improvements in metrics like Cautious Response and False Resistance were not observed, as the model did not respond cautiously when uncertain, contrary to our initial assumptions. Instead, abstention led to results that were largely unremarkable. The accuracy remained comparable to the baseline, and Confidence Balance showed minimal improvement, or in some instances, performed worse than the zero-shot setup, as seen in Tab. \ref{fig:heatmap}.

% \begin{figure}[t]

%   \includegraphics[width=1\columnwidth]{latex/performance_comparison (9).pdf}
%   \caption{Accuracy distribution for all the models across different prompts.}
%   \label{fig:CWAM}
% \end{figure}

\subsection{Confidence Balance Variability}
\begin{figure}[t]

  \includegraphics[width=1\columnwidth]{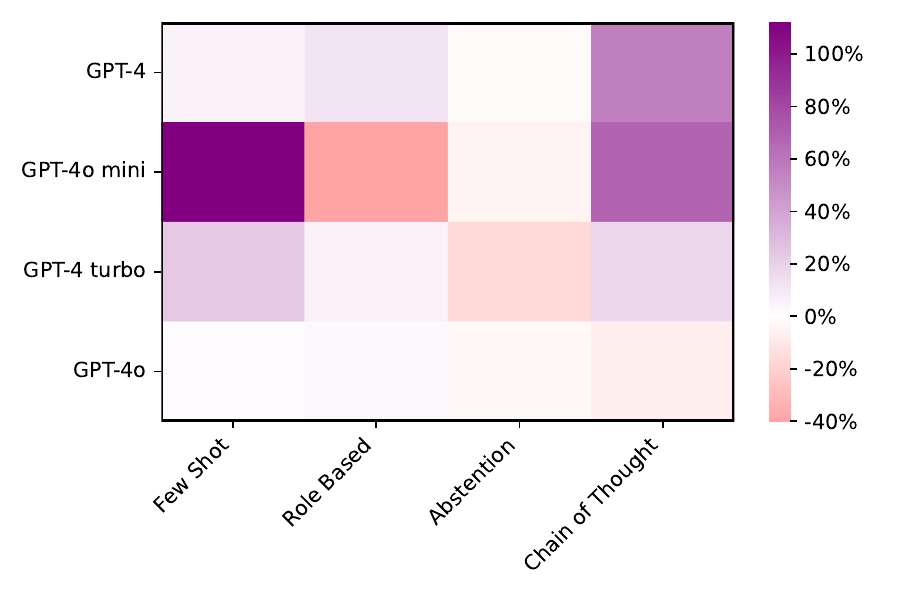}
  \caption{Heatmap visualization of models performance of different prompts on Confidence Balance with respect to zero-shot prompt.}
  \label{fig:heatmap}
\end{figure}

While the overall accuracy did not improve significantly, we observed some fluctuations compared to the zero-shot approach for CB (Confidence Balance). This metric integrates both confidence and accuracy, providing insight into how each model evaluates the answers it generates. Even though the accuracy for few-shot setting was lower, CB score of the the models improved compared to zero-shot which is depicted in Fig. \ref{fig:heatmap}. The two examples provided in the prompt helped guide the models toward generating responses with greater confidence in the correct answers. This suggests that when models are confident in their responses, they are less prone to confusion or generating conflicting answers. Similarly, for CoT prompting, breaking down problems into smaller subproblems and analyzing them individually enhances the model’s ability to provide confident and accurate answers.

\subsection{Hallucination for unanswerable question}
As illustrated in Fig. \ref{fig:comparison}, GPT models often exhibit hallucinations when they fail to fully comprehend the question or misinterpret certain words or phrases, leading to incorrect answers. Our step-by-step analysis during CoT prompting revealed that the models sometimes fabricate questions that were not originally present when breaking down the problem to solve it. While GPT-4’s performance with CoT prompting surpasses that of other prompts, it still struggles with generating incorrect answers for unanswerable questions. Table \ref{tab:question_cot} illustrates the model’s reasoning process for these unanswerable questions, showcasing the step-by-step approach taken by the model across different question categories in the dataset.

\begin{table*}[ht]
\centering
\small  % Set the font size for the entire table
\begin{tabular}{|M{1.5cm}|M{6cm}|M{6cm}|}

\hline
\textbf{Category} & \textbf{Unanswerable Question} & \textbf{CoT Explanation} \\
\hline

Missing Key Info & Paige had 43 math problems and 12 science problems for homework. If she finished 44 of the math and science problems at school, how many problems did she have to do for \textcolor{red}{math} homework?
&

Paige had 43 math problems and 12 science problems, making a total of 43 + 12 = 55 problems. If she finished 44 of the math and science problems at school, she has 55 - 44 = 11 problems left to do for math homework. \\

\hline

Ambiguous Key Info & For the school bake sale Bianca made \textcolor{red}{more than} 14 cupcakes. If she sold 6 of them and then made 17 more, how many cupcakes would she have? 

&

Bianca made more than 14 cupcakes. She sold 6 of them and then made 17 more. So, the total number of cupcakes she would have is (14+17)-6.\\

\hline

Unrealistic & Jerry had some action figures on a shelf in his room. Later he added 2 more action figures to the shelf and removed 7 of the old ones. If there are 10 action figures on his shelf now.How many \textcolor{red}{anime} figures did he have initially on the shelf?
& 
Jerry added 2 more action figures to the shelf and removed 7 of the old ones. If there are 10 action figures on his shelf now, to find the initial number of action figures, we add the number of action figures removed (7) to the final number of action figures (10) and subtract the number of action figures added (2). \\
\hline

Unrelated & Rachel had to complete 2 pages of reading homework and 4 pages of math homework. How many more pages of \textcolor{red}{speaking} homework than \textcolor{red}{reading} homework did she have?
&
Rachel had to complete 2 pages of reading homework and 4 pages of math homework. So, we subtract the number of pages of reading homework from the number of pages of math homework to find out how many more pages of speaking homework than reading homework she had.
4 (math homework) - 2 (reading homework) = 2\\

\hline

Incomplete Ques & The grasshopper and the frog had a jumping contest. The grasshopper jumped 31 inches and the frog jumped 35 inches.
& 
The grasshopper jumped 31 inches and the frog jumped 35 inches. So, we add the distances jumped by the grasshopper and the frog to find out the total distance jumped. 31 (grasshopper's jump) + 35 (frog's jump) = 66\\
\hline

\end{tabular}
\caption{The table displays CoT analysis of GPT-4 for the unanswerable questions for each category. The sections highlighted in \textcolor{red}{red} denote the elements that contribute to the question being unanswerable.}
\label{tab:question_cot}
\end{table*}

The \textbf{Missing Key Info} category, as the name suggests, contains questions with missing key information. In the given question, since the number of completed problems is not provided, it becomes impossible to determine how many math problems remain. The model incorrectly assumed that all remaining problems are math-related, overlooking the possibility of science problems. The \textbf{Unrealistic} and \textbf{Unrelated} categories exhibited similar issues, where the model fails to comprehend the changes or perturbations in the questions but attempts to answer them regardless. In \textbf{Ambiguous Key Info}, additional information introduced ambiguity, yet the model fails to identify this ambiguity. Finally, in \textbf{Incomplete Ques}, the model proceeded to answer without a properly framed question. We can draw several key conclusions from our observations. Remarkably, the model attempts to generate answers even when posed with unanswerable questions. What is surprising is that it often provides the correct response from the set of answerable questions. Specifically, for GPT-4 with Chain of Thought (CoT), 77\% of the answers given for unanswerable questions corresponded to the correct answers from answerable ones. This suggests that the model tries to make sense of the question, despite its ambiguity or unanswerable nature. It either disregards the uncertainty or reimagines the question logically based on the provided information. This behavior indicates that the model might be reconstructing or correcting the question to fit the scenario and generate a plausible answer as seen from Tab. \ref{tab:question_cot} where it corrected itself to "action figures" instead of "anime figures" and proceeded to analyze.

\subsection{Analysis of Unanswerable Question Categories}

\begin{figure*}[t]
  \centering % Adjust the value to move the image to the right
  \includegraphics[width=\textwidth]{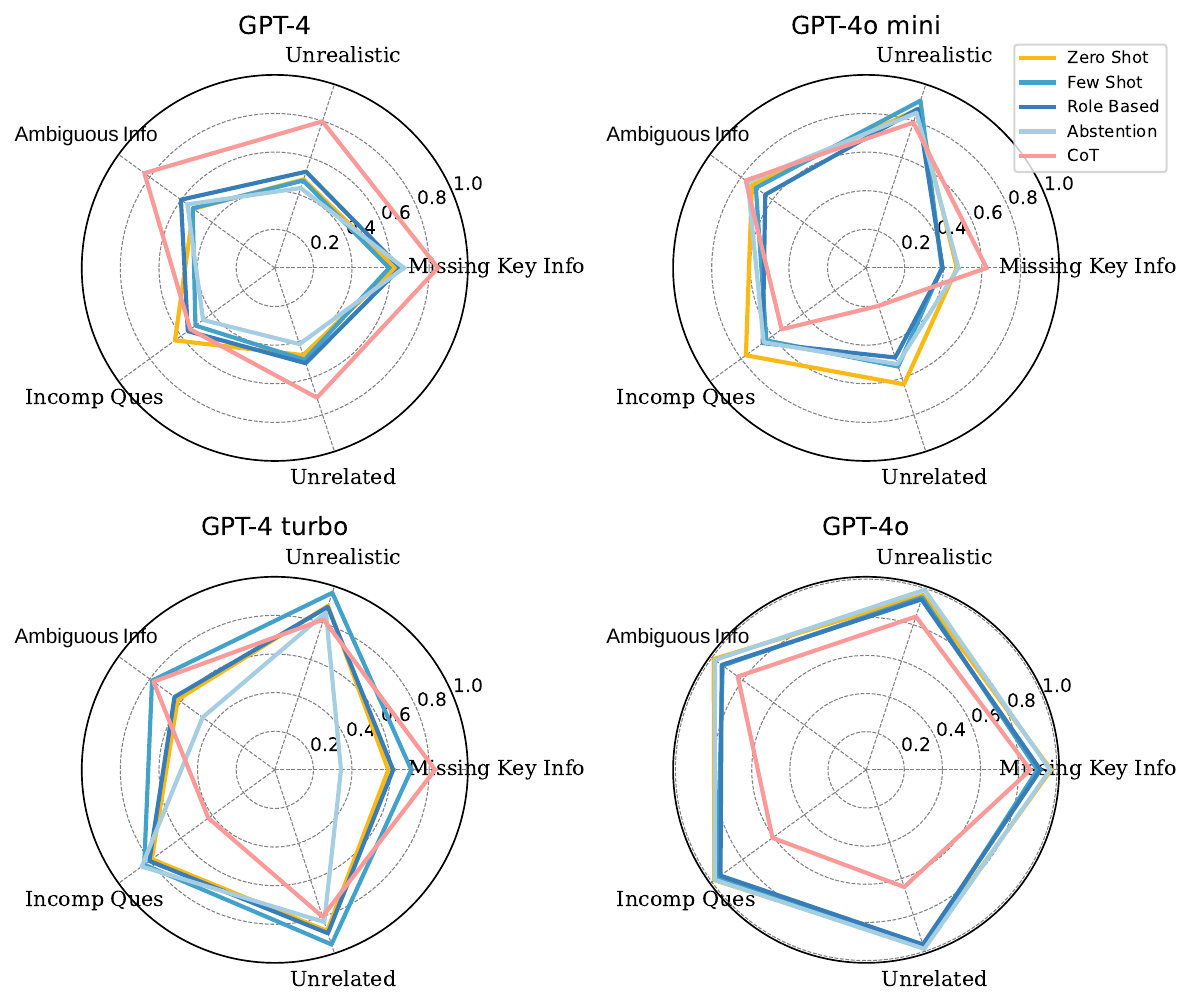}
  \caption{ Radar chart depicting the performance of each model across various categories of unanswerable questions, evaluated under different prompt strategies. This visualization highlights the optimal model-prompt combinations for handling specific types of unanswerable scenarios. }
  \label{fig:question_cat}
\end{figure*}

Fig. \ref{fig:question_cat} illustrates the impressive performance of GPT-4o, which exhibits near-perfect results across various question types, with the exception of CoT prompts, where there is a significant performance drop. Interestingly, while CoT negatively impacts GPT-4o, it enhances the performance of all other models, particularly GPT-4, which struggles with alternative prompt types.

Notably, CoT tends to perform poorly with incomplete questions, as it attempts to address problems incrementally, failing to recognize their unanswerable nature. The data for GPT-4o mini reveals that it ranks lowest among the models.

Conversely, the zero-shot approach shows a commendable ability to identify when to select NOTA/IDK responses, effectively indicating when a question is unsolvable. Furthermore, the few-shot, role-based (math expert) and abstention strategies, yield results comparable to those of the zero-shot model.

In summary, while CoT can be detrimental in scenarios involving incomplete questions, it generally improves performance in other contexts by aiding models in discerning when to refrain from answering. Overall, GPT-4o stands out as the most effective model, though GPT-4 turbo also demonstrates a similar proficiency in recognizing unanswerable questions under certain prompts. 

\section{Discussion}

This study aims to analyze which GPT model performs best on unanswerable questions when used in combination with different prompts. The prompts neither significantly improved overall accuracy nor influenced the model's abstention behavior when faced with unanswerable questions. CoT reasoning improved GPT-4's performance in certain categories of questions that is similar to GPT-4o. Interestingly, a zero-shot or base prompt often performed well, while few-shot, role-based, or abstention-specific prompts are less effective. This can be attributed to GPT's tendency to hallucinate and force an answer, attempting to provide a plausible response even for unanswerable questions. In essence, GPT models prioritize making a question answerable rather than selecting options like IDK or NOTA.

\section{Conclusion}
Given the potential for GPT models to be used as tools for solving mathematical problems in the near future, the ability to distinguish unanswerable questions becomes a critical feature. Our objective was to evaluate the performance of GPT models with various prompts using novel metrics we developed. We demonstrated how these models often hallucinate to unanswerable questions, even when the option to abstain is available. Our findings show that advanced prompts do not significantly improve this behavior, highlighting the need for models to better recognize when to abstain from answering or accurately identify issues in the question.

\section{Limitations}

We were unable to evaluate models of o1 series from OpenAI which are among the most recent and high-performing versions, due to access restrictions. Additionally, we did not explore the niche prompts commonly employed in other studies on large language models (LLMs). Another limitation lies in the dataset itself: the math word problems we used were not highly complex, and we did not assess model performance across varying levels of difficulty. Our evaluation focused solely on word problems, without extending to other mathematical categories such as algebra or geometry.

\bibliography{custom}

\end{document}